\documentclass[11pt,a4paper]{article}
\usepackage[hyperref]{acl2020}
\usepackage{times}
\usepackage{latexsym}

\usepackage{microtype}
\usepackage{xspace}
\usepackage{xcolor}
\usepackage{enumitem}
\usepackage{bm}
\usepackage{graphicx}
\usepackage{subcaption}
\usepackage{amsmath,amsfonts,amssymb}
\usepackage{url}
\usepackage{array,multirow}

\usepackage{adjustbox}

\newcommand{\modelname}{\texttt{PCPR}\xspace}
\newcommand{\simplemodelname}{\texttt{CPR}\xspace}
\newcommand{\modelfullname}{\underline{P}ronunciation-attentive \underline{C}ontextualized \underline{P}un \underline{R}ecognition\xspace}

\DeclareMathOperator*{\argmax}{argmax}

\aclfinalcopy 

\title{``The Boating Store Had Its Best Sail Ever'':\\ Pronunciation-attentive Contextualized Pun Recognition}

\author{Yichao Zhou, Jyun-yu Jiang, Jieyu Zhao, Kai-Wei Chang and Wei Wang \\
    Computer Science Department\\
    University of California, Los Angeles \\
  \texttt{\{yz, jyunyu, jyzhao, kwchang, weiwang\}@cs.ucla.edu}}

\date{}

\begin{document}
\maketitle
\begin{abstract}
Humor plays an important role in human languages and it is essential to model humor when building intelligence systems. Among different forms of humor,  
puns perform wordplay for humorous effects by employing words with double entendre and high phonetic similarity.
However, identifying and modeling puns are challenging as puns usually involved implicit semantic or phonological tricks.
In this paper, we propose \modelfullname~(\modelname) to perceive human humor, detect if a sentence contains puns and locate them in the sentence. 
\modelname derives contextualized representation for each word in a sentence by capturing the association between the surrounding context and its corresponding phonetic symbols.
Extensive experiments are conducted on two benchmark datasets. Results demonstrate that the proposed approach significantly outperforms the state-of-the-art methods  
in pun detection and location tasks. 
In-depth analyses verify the effectiveness and robustness of \modelname.

\end{abstract}

\section{Introduction}
\label{section:introduction}
During the last decades, social media has promoted the creation of a vast amount of humorous web contents~\cite{nijholt2017humor}. 
Automatic recognition of humor has become an important task in the area of figurative language processing, which can benefit various downstream NLP applications such as dialogue systems, sentiment analysis, and machine translation~\cite{melby1995possibility,augello2008humorist,ghosh2015semeval,bertero2016predicting,blinov2019large}. 
However, humor is one of the most complicated behaviors in natural language semantics and sometimes it is even difficult for humans to interpret. In most cases, understanding humor requires adequate background knowledge and a rich context.

Puns are a form of humorous approaches using the different meanings of identical words or words with similar pronunciations to explain texts or utterances.
There are two main types of puns. Homographic puns rely on multiple interpretations of the same word. As shown in Table~\ref{tab:puns}, the phrase \emph{all right} means \emph{good condition} or \emph{opposite to left}; the word \emph{reaction} means \emph{chemical change} or \emph{action}. 
The two meanings of the same expression are consistent with its context, which creates a humorous pun in both sentences when there is a clear contrast between two meanings.
On the other hand, heterographic puns take advantage of phonologically same or similar words. For example, the word pairs \emph{sale} and \emph{sail}, \emph{weak} and \emph{week} in Table~\ref{tab:puns} have the same or similar pronunciations. The sentences are funny because both words fit the same context. 
Understanding puns is a big fish to fry for deep comprehension of complex semantics.

\begin{table}[t!]
    \centering
    \resizebox{\linewidth}{!}{
    \begin{tabular}{|ll|}
    \hline
    & \textbf{Homographic Puns}\\
    \hline
    1. & Did you hear about the guy whose whole left side was cut\\  
    & off? He's {\color{red}\textbf{all right}} now. \\
    2. & I'd tell you a chemistry joke but I know I wouldn't get a \\ 
    & {\color{red}\textbf{reaction}}. \\
    \hline
    \hline
    &\textbf{Heterographic Puns}\\
    \hline
    1. & The boating store had its best {\color{red}\textbf{sail (sale)}} ever. \\ 
    2. & I lift weights only on Saturday and Sunday because Monday\\ 
    & to Friday are {\color{red}\textbf{weak (week)}} days.\\
    \hline
    \end{tabular}}
    \caption{Examples of homographic and heterographic puns.}
    \label{tab:puns}
\end{table}

These two forms of puns have been studied in literature from different angles. To recognize puns in a sentence, word sense disambiguation techniques~(WSD)~\cite{navigli2009word} have been employed to identify the equitable intention of words in utterances~\cite{pedersen2017duluth}.
External knowledge bases such as {WordNet}~\cite{miller1998wordnet} have been applied in determining word senses of pun words~\cite{oele2017buzzsaw}.
However, these methods cannot tackle heterographic puns with distinct word spellings and knowledge bases that only contain a limited vocabulary.
To resolve the issues of sparseness and heterographics, the word embedding techniques~\cite{mikolov2013distributed,pennington2014glove} provide flexible representations to model puns~\cite{hurtado2017elirf,indurthi2017fermi,cai2018sense}.
However, a word may have different meanings regarding its contexts.
Especially, an infrequent meaning of the word might be utilized for creating a pun. 
Therefore, static word embeddings are insufficient to represent words. 
In addition, some puns are created by 
replacing a word with another word with the same or similar pronunciation as examples shown in Table~\ref{tab:puns}. 
Therefore, to recognize puns, it is essential to model the association between words in the sentence and the pronunciation of words. Despite existing approaches attempt to leverage phonological structures to understand puns~\cite{doogan2017idiom,jaech2016phonological}, there is a lack of a general framework to model these two types of signals in a whole. 



In this paper, we propose \modelfullname~(\modelname) to jointly model the contextualized word embeddings and phonological word representations for pun recognition. 
To capture the phonological structures of words, we break each word into a sequence of phonemes as its pronunciation so that homophones can have similar phoneme sets.
For instance, the phonemes of the word \emph{pun} are \{\texttt{P}, \texttt{AH}, \texttt{N}\}.
In \modelname, we construct a pronunciation attentive module to identify important phonemes of each word, which can be applied in other tasks related to phonology. 
We jointly encode the contextual and phonological features into a self-attentive embedding to tackle both pun detection and location tasks.
We summarize our contributions as following.
\begin{itemize}[leftmargin=*,itemsep=0pt]
    \item To the best of our knowledge, \modelname is the first work to jointly model contextualized word embeddings and pronunciation embeddings to recognize puns. Both contexts and phonological properties are beneficial to pun recognition.
    \item Extensive experiments are conducted on two benchmark datasets. 
    \modelname significantly outperforms existing methods in both pun detection and pun location.
    In-depth analyses also verify the effectiveness and robustness of \modelname.
    \item We release our implementations and pre-trained phoneme embeddings at \url{https://github.com/joey1993/pun-recognition} to facilitate future research.
\end{itemize}
\section{Related Work}
\label{section:relatedwork}
\noindent\textbf{{Pun Recognition and Generation}}
To recognize puns, \citet{miller2017semeval} summarize several systems for the SemEval 2017 tasks. 
To detect the pun, \citet{pedersen2017duluth} supposes that if there is one pun in the sentence, when adopting different Word Sense Disambiguation (WSD) methods, the sense assigned to the sentence will be different. To locate the pun, based on the WSD results for pun detection, they choose the last word which changes the senses between  different WSD runs. Even though this method can tackle both homographic and heterographic pun detection, it does not use any pre-trained embedding model.
\citet{xiu2017ecnu} detect the pun in the sentence using similarity features which are calculated on sense vectors or cluster center vectors. To locate the pun, they use an unsupervised system by scoring each word in the sentence and choosing the word with the smallest score. However, this model exclusively relies on semantics to detect the heterographic puns but ignores the rich information embedded in the pronunciations.
\citet{doogan2017idiom} leverage word embeddings as well as the phonetic information by concatenating pronunciation strings, but the concatenation has limited expression ability. They also mention that their systems suffer for short sentences as word embeddings do not have much context information.

Besides, \citet{zoujoint} jointly detect and locate the pun from  a sequence labeling perspective by employing a new tagging schema.  
\citet{diao2018weca} expand word embeddings using WordNet to settle the polysemy of homographic puns, following by a  neural attention mechanism to extract the collocation to detect the homographic pun. However, all these methods only make use of limited context information. 
Other than the pun recognition, \citet{yu2018a} generate homographic puns without requiring any pun data for training. \citet{he2019pun} improve the homographic pun generation based on the ``local-global surprisal principle'' which posits that the pun word and the alternative word have a strong association with the distant and immediate context respectively.  

\noindent \textbf{Pronunciation Embeddings}
Word embeddings assign each word with a vector so that words with similar semantic meanings are close in the embedding space.  Most word embedding models   only make use of  text information and omitting the rich information contained in the pronunciation. However, the pronunciation is also an important part of the language~\cite{zhu2018improve}. Prior studies have demonstrated that the phonetic information can be used in speech recognition~\cite{bengio2014word}, spell correction~\cite{toutanova2002pronunciation} and  speech synthesis~\cite{miller1998pronunciation}. By projecting to the embedding space, words sound alike are nearby to each other~\cite{bengio2014word}.  
Furthermore, \citet{kamper2016deep} make use of word pairs information to improve the acoustic word embedding. \citet{zhu2018improve} show that combining the pronunciation with the writing texts can help to improve the performance of word embeddings. However, these pronunciation embeddings are word-level features, while in our approach, we make use of syllabic pronunciations which is phoneme-level and could help with the out-of-vocabulary (OOV) situation. \citet{luo-etal-2019-pun} also propose an adversarial generative network for pun generation, which does not require any pun corpus.

\noindent \textbf{{Contextualized Word Embeddings}}
Traditional word embeddings assign a fixed vector to one word even if the word has multiple meanings under different contexts (e.g., ``the river bank'' v.s. ``the commercial bank'').
\citet{mccann2017learned} combine the pivot word embeddings as well as  the contextual embeddings generated by an encoder from a supervised neural machine translation task. \citet{peters2017semi} enrich the word embeddings by the contextual information extracted from a bi-directional language model. 
\cite{devlin2018bert} learn the language embedding by stacking multiple transformer layers with masked language model objective which advances the state-of-the-art for many NLP tasks. 
\citet{yang2019xlnet} enable learning bidirectional contexts by maximizing the expected likelihood over all permutations of the factorization order and solve the problem of pretrain-finetune discrepancy.

\section{Pronunciation-attentive \\Contextualized  Pun Recognition}
\label{section:method}

In this section, we first formally define the problem and then introduce the proposed method, \modelname.

\subsection{Problem Statement}
\label{section:probstate}

Suppose the input text consists of a sequence of $N$ words  $\left\lbrace w_1, w_2, \cdots, w_N \right\rbrace$.
For each word $w_i$ with $M_i$ phonemes in its pronunciation, the phonemes are denoted as $R(w_i) = \left\lbrace r_{i,1}, r_{i,2}, \cdots, r_{i, M_i} \right\rbrace $, where $r_{i,j}$ is the $j$-th phoneme in the pronunciation of $w_i$. These phonemes are given by a dictionary.
In this paper, we aim to recognize potential puns in the text with two tasks, including pun detection and pun location, as described in the following.

\noindent \textbf{Task 1: Pun Detection.}
The pun detection task identifies whether a sentence contains a pun. Formally, the task is modeled as a classification problem with binary label $y^D$. 

\noindent \textbf{Task 2: Pun Location.}
Given a sentence containing at least a pun, the pun location task aims to unearth the pun word.
More precisely, for each word $w_i$, we would like to predict a binary label $y^L_i$ that indicates if $w_i$ is a pun word.

In addition to independently solving the above two tasks, the ultimate goal of pun recognition is to build a pipeline from scratch to detect and then locate the puns in texts.
Hence, we also evaluate the end-to-end performance by aggregating the solutions for two tasks.

\begin{figure*}[!t]
    \centering
    \includegraphics[width=.85\linewidth]{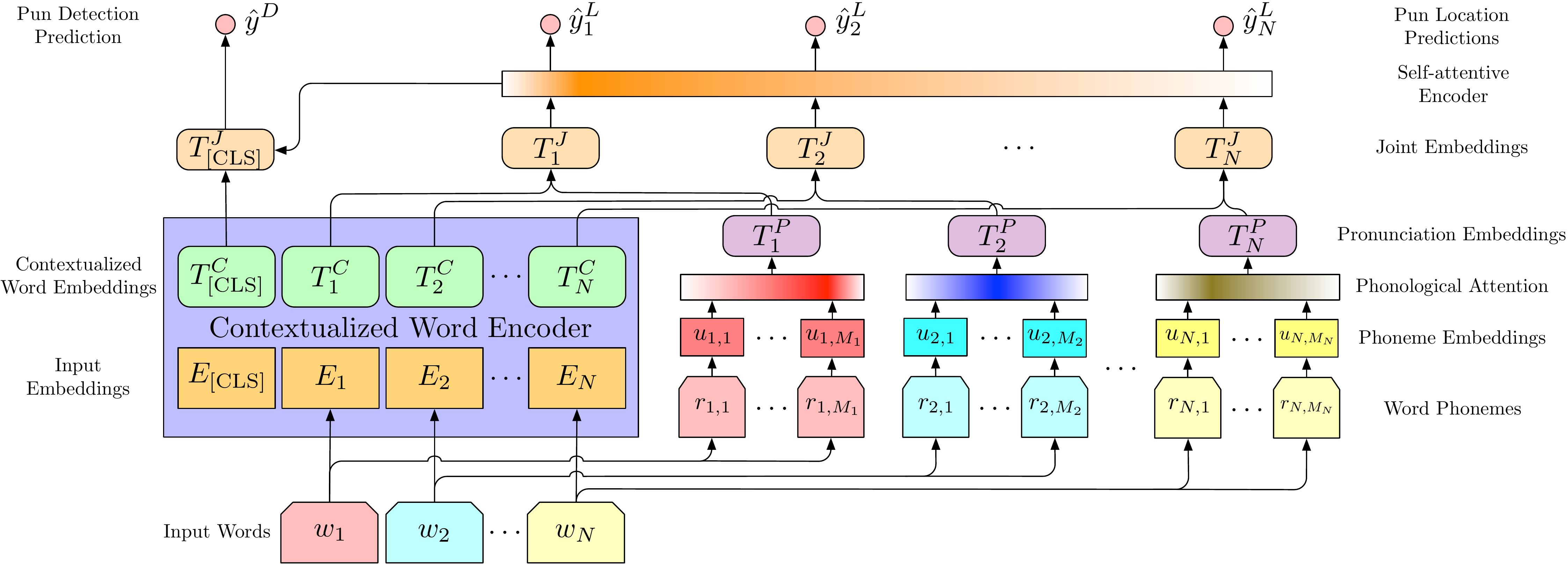}
    \caption{The overall framework of \modelname. We leverage the self-attention mechanism to jointly model contextualized embeddings and phonological representations. \modelname can tackle both pun detection and pun location tasks.}
    \label{fig:framework}
\end{figure*}

\subsection{Framework Overview}

Figure~\ref{fig:framework} shows the overall framework of the proposed \modelfullname~(\modelname).
For each word in the input text, we first derive two continuous vectors, including contextualized word embedding and pronunciation embedding, as representations in different aspects.
Contextualized word embeddings derive appropriate word representations with consideration of context words and capture the accurate semantics in the text.
To learn the phonological characteristics, each word is divided into phonemes while each phoneme is projected to a phoneme embedding space, thereby obtaining pronunciation embeddings with the attention mechanism~\cite{bahdanau2014neural}.
Finally, a self-attentive encoder blends contextualized word embeddings and pronunciation embeddings to capture the overall semantics for both pun detection and location.

\subsection{Contextualized Word Embeddings}

The context is essential for interpreting a word in the text.
Hence, we propose to apply contextualized word embeddings to derive word representations.
In the framework of \modelname, any contextualized word embedding method, such as BERT~\cite{devlin2018bert}, ELMo~\cite{peters2018deep}, and XLNet~\cite{yang2019xlnet}, can be utilized.
Here, we choose BERT to derive contextualized word embeddings without loss of generality.

BERT deploys a multi-layer bidirectional encoder based on transformers with multi-head self-attention~\cite{vaswani2017attention} to model words in the text after integrating both word and position embeddings~\cite{sukhbaatar2015end}.
As a result, for each word, a representative contextualized embedding is derived by considering both the specific word and all contexts in the document.
Here we denote $T^C_i$ as the $d_C$-dimensional contextualized word embedding for the word $w_i$.
In addition, BERT contains a special token \texttt{[CLS]} with an embedding vector in BERT to represent the semantics of the whole input text.

\subsection{Pronunciation Embeddings}

To learn the phonological characteristics of words, \modelname models the word phonemes.
For each phoneme $r_{i, j}$ of the word $w_i$, we project $r_{i, j}$ to a $d_P$-dimensional embedding space as a trainable vector $u_{i, j}$ to represent its phonological properties.

Based on the phoneme embeddings of a word, we apply the attention mechanism~\cite{bahdanau2014neural} to simultaneously identify important phonemes and derive the pronunciation embedding $T^P_i$.
Specifically, the phoneme embeddings are transformed by a fully-connected hidden layer to measure the importance scores $\alpha^P_i$ as follows:
$$v_{i, j} =  \tanh(\mathcal{F}_P(u_{i,j})),$$
$$\alpha^P_{i,j} = \frac{v_{i,j}^\intercal v_s}{\sum_k v_{i,k}^\intercal v_s },$$
where $\mathcal{F}_P(\cdot)$ is a fully-connected layer with $d_A$ outputs and $d_A$ is the attention size; $v_s$ is a $d_A$-dimensional context vector that estimates the importance score of each pronunciation embedding.
Finally, the pronunciation embeddings $T^P_i$ can be represented as the weighted combination of phoneme embeddings as follows:
$$T^P_i = \sum_j \alpha_{i,j} u_{i,j}.$$ 
Moreover, we can further derive the joint embedding $T^J_i$ to indicate both word semantics and phonological knowledge for the word $w_i$ by concatenating two different embeddings as follows:
$$T^J_i = \left[T^C_i; T^P_i\right].$$
Note that the joint embeddings are $d_J$-dimensional vectors, where $d_J = d_C + d_P$.

\subsection{Pronunciation-attentive Contextualized Embedding with Self-attention}

For the task of pun detection,  understanding the meaning of input text is essential.
Due to its advantages of interpretability over convolutional neural network~\cite{lecun1995convolutional} and recurrent neural network~\cite{schuster1997bidirectional}, 
we deploy the self-attention mechanism~\cite{vaswani2017attention} to capture the overall semantics represented in the joint embeddings.
For each word $w_i$, the self-attention mechanism estimates an importance vector $\alpha^S_i$:
\vspace{-10pt}
$$\mathcal{F}_S(T) = \text{Softmax}(\frac{TT^\intercal}{\sqrt{d}})T,$$ 
$$\alpha^S_i = \frac{\exp(\mathcal{F}_S(T^J_i))}{\sum_j \exp(\mathcal{F}_S(T^J_j))},$$
where $\mathcal{F}_S(\cdot)$ is the function to estimate the attention for queries, and  $d$ is a  scaling factor to avoid extremely small gradients. 
Hence, the self-attentive embedding vector is computed by aggregating joint embeddings:
$$T^J_{\texttt{[ATT]}} = \sum_i \alpha^S_i \cdot T^J_i.$$ 
Note that the knowledge of pronunciations is considered by the self-attentive encoder but not the contextualized word encoder.
Finally, the pronunciation-attentive contextualized representation for the whole input text can be derived by concatenating the overall contextualized embedding and the self-attentive embedding:
$$T^J_{\texttt{[CLS]}} = \left[T^C_{\texttt{[CLS]}} ;T^J_{\texttt{[ATT]}}\right].$$
Moreover, each word $w_i$ is benefited from the self-attentive encoder and 
is represented by a joint embedding:
$$T^J_{\texttt{i,[ATT]}} = \alpha^S_i \cdot T^J_i.$$

\subsection{Inference and Optimization}

Based on the joint embedding for each word and the pronunciation-attentive contextualized embedding for the whole input text, both tasks can be tackled with simple fully-connected layers.

\noindent \textbf{Pun Detection.}
Pun detection is modeled as a binary classification task.
Given the overall embedding for the input text $T^J_{\texttt{[CLS]}}$, the prediction $\hat{y}^D$ is generated by a fully-connected layer and the softmax function:
$$\hat{y}^D = \argmax_{k\in\lbrace 0,1 \rbrace }{\mathcal{F}_D(T^J_{\texttt{[CLS]}})_k},$$
where $\mathcal{F}_D(\cdot)$ derives the logits of two classes in binary classification.

\noindent \textbf{Pun Location.}
For each word $w_i$, the corresponding self-attentive joint embedding $T^J_{\texttt{i,[ATT]}}$ is applied as features for pun location.
Similar to pun detection,  the prediction $\hat{y}^L_i$ is generated by:
$$\hat{y}^L_i = \argmax_{k\in\lbrace 0,1 \rbrace }{\mathcal{F}_L(T^J_{\texttt{i,[ATT]}})_k},$$
where $\mathcal{F}_L(\cdot)$ derives two logits for classifying if a word is a pun word.

Since both tasks focus on binary classification, we optimize the model with cross-entropy loss. 

\section{Experiments}
\label{section:experiments}
In this section, we describe our experimental settings and explain the results and interpretations. We will verify some basic assumptions of this paper: (1) the contextualized word embeddings and pronunciation embeddings are both beneficial to the pun detection and location tasks; (2) the attention mechanism can improve the performance. 

\subsection{Experiment settings}

\begin{table}[!t]
    \centering
    \resizebox{.9\linewidth}{!}{
    \begin{tabular}{|c|cc|c|}\hline
     \multirow{2}{*}{Dataset} & \multicolumn{2}{c|}{SemEval} & \multirow{2}{*}{PTD}  \\
             & Homo & Hetero &  \\ \hline
    Examples w/ Puns & 1,607 & 1,271 & 2,423 \\
    Examples w/o Puns & 643 & 509 & 2,403 \\ \hline
    Total Examples & 2,250 & 1,780 & 4,826 \\ \hline
    \end{tabular}}
    \caption{Data statistics.  ``Homo'' and ``Hetero'' denote homographic and heterographic puns. Pun detection employs all of the examples in the two datasets while pun location only exploits the examples with puns in SemEval due to the limitation of annotations.}
    \label{tab:data}
\end{table}

\noindent \textbf{Experimental Datasets.}
We conducted experiments on the SemEval 2017 shared task~7 dataset\footnote{\url{http://alt.qcri.org/semeval2017/task7/}}~(SemEval)~\cite{miller2017semeval} and the Pun of The Day dataset~(PTD)~\cite{yang2015humor}.
For pun detection, the SemEval dataset consists of $4,030$  and $2,878$ examples for pun detection and location while each example with a pun can be a homographic or heterographic pun.
In contrast, the PTD dataset contains $4,826$  examples without labels of pun types.
Table~\ref{tab:data} further shows the data statistics. The two experimental datasets are the largest publicly available benchmarks  that are used in the existing studies. 
SemEval-2017 dataset contains punning and non-punning jokes, aphorisms, and other short texts composed by professional humorists and online collections. Hence, we assume the genres of positive and negative examples should be identical or extremely similar.


\noindent \textbf{Evaluation Metrics.}
We adopt precision~(P), recall~(R), and \textit{F}$_{1}$-score~\cite{schutze2007introduction,powers2011evaluation} to compare the performance of \modelname with previous studies in both pun detection and location. More specifically, we apply 10-fold cross-validation to conduct evaluation. For each fold, we randomly select 10\% of the instances from the training set for development.
To conduct fair comparisons, we strictly follow the experimental settings in previous studies~\cite{zoujoint,cai2018sense} and include their reported numbers in the comparisons.  

\noindent \textbf{Implementation Details.}
For data pre-processing, all of the numbers and punctuation marks are removed.
The phonemes of each word are derived by the CMU Pronouncing Dictionary\footnote{\url{http://svn.code.sf.net/p/cmusphinx/code/trunk/cmudict/}}.
We initialize the phoneme embeddings by using the \emph{fastText} word embedding~\cite{mikolov2018advances} trained on Wikipedia articles\footnote{\url{https://dumps.wikimedia.org/enwiki/latest/}} crawled in December, 2017.
The \modelname is implemented in PyTorch while the fused Adam optimizer~\cite{kingma2014adam} optimizes the parameters with an initial learning rate of $5\times 10^{-5}$.
The dropout and batch size are set as $10^{-1}$ and 32.
We follow BERT~(BASE)~\cite{devlin2018bert} to use 12 Transformer layers and self-attention heads. To clarify, in \modelname, tokens and phonemes are independently processed, so the tokens processed with WordPiece tokenizer~\cite{wu2016google} in BERT are not required to line up with phonemes for computations. To deal with the out-of-vocabulary words, we use the output embeddings of the first WordPiece tokens as the representatives, which is consistent with many  state-of-the-art named entity recognition approaches~\cite{devlin2018bert,lee2019biobert}.
We also create a variant of \modelname called \texttt{CPR} by exploiting only the contextualized word encoder without considering phonemes to demonstrate the effectiveness of pronunciation embeddings.

\begin{figure}[!t]
\centering
    \begin{subfigure}[b]{0.49\linewidth}
        \includegraphics[width=1\linewidth]{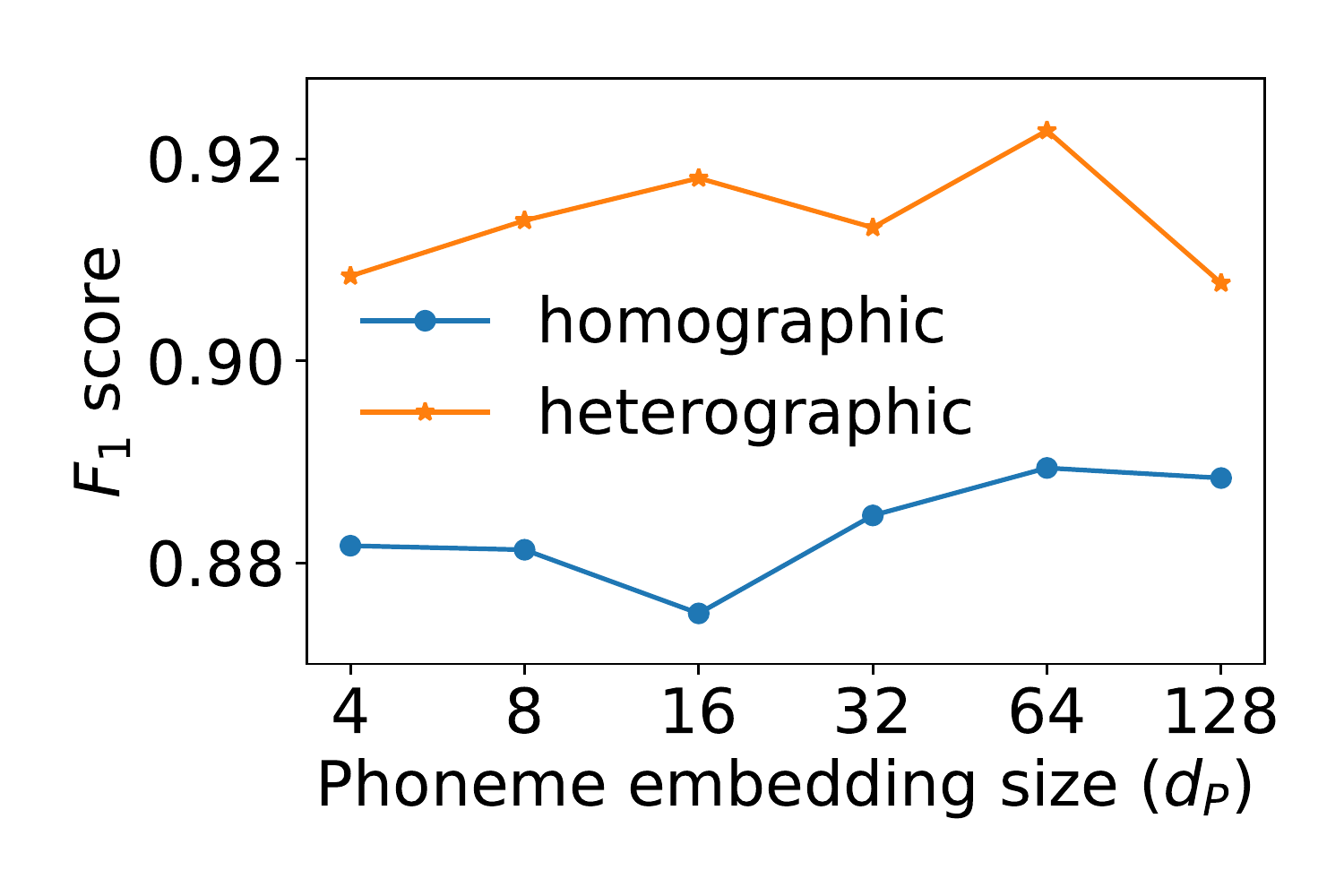}
        \caption{Phoneme emb. size $d_P$}
        \label{fig:phoneme}
    \end{subfigure}
    \begin{subfigure}[b]{0.49\linewidth}
        \includegraphics[width=1\linewidth]{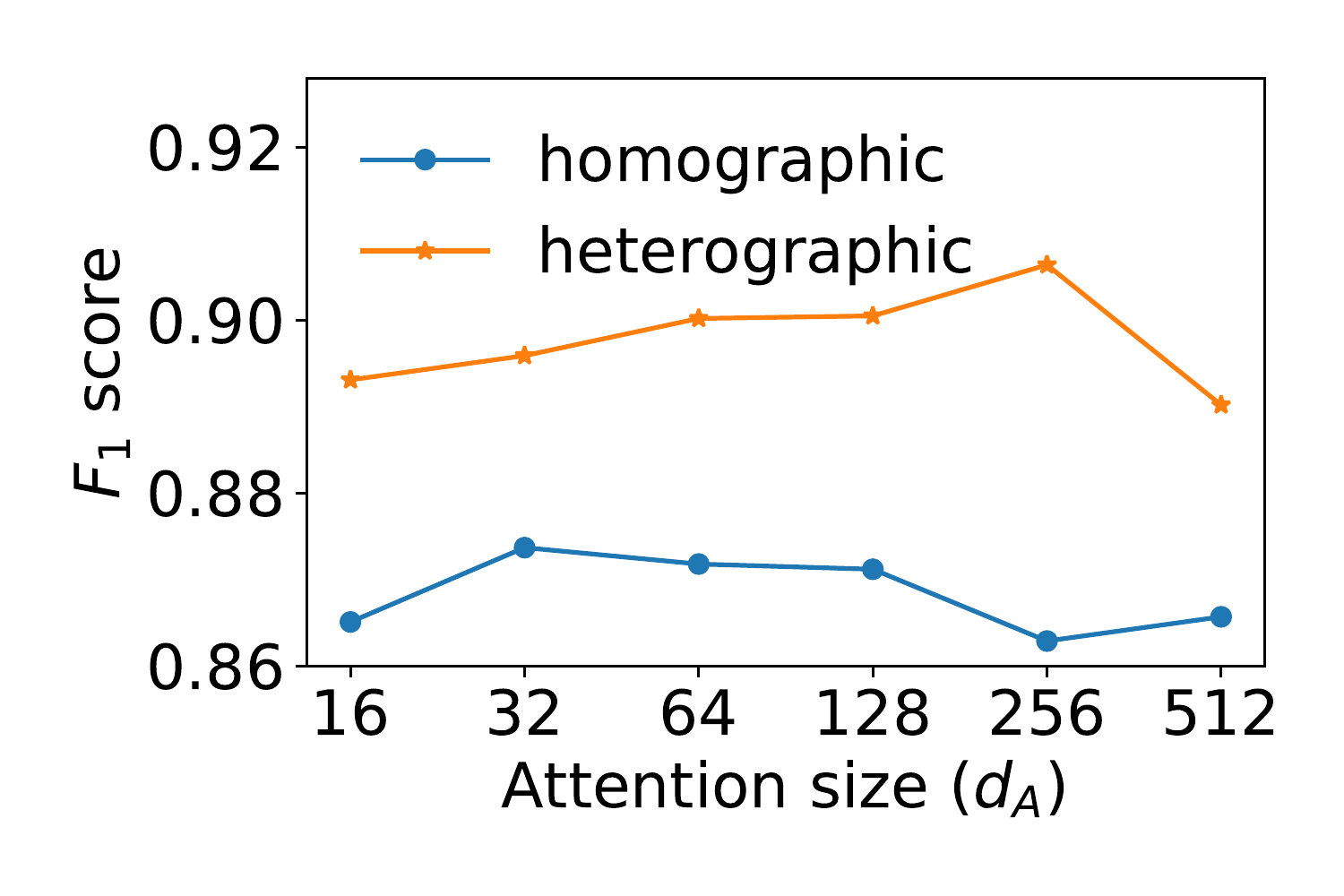}
        \caption{Attention size $d_A$}
        \label{fig:hidden}
    \end{subfigure}
    \caption{
    Pun location performance over different phoneme embedding sizes $d_P$ and attention sizes $d_A$ on the SemEval dataset.
    }
    \label{fig:dimension}
\end{figure}

To tune the hyperparameters, we search the phoneme embedding size $d_P$ and the attention size $d_A$ from $\lbrace 8, 16, 32, 64, 128, 256, 512 \rbrace$ as shown in Figure~\ref{fig:dimension}.
For the SemEval dataset, the best setting is $(d_P=64, d_A=256)$ for the homographic puns while heterographic puns favor $(d_P=64, d_A=32)$.
For the PTD dataset, $(d_P=64, d_A=32)$ can reach the best performance.

\noindent \textbf{Baseline Methods.} 
We compare \modelname with several baseline methods.

For the SemEval dataset, nine baseline methods are compared in the experiments, including \texttt{Duluth}~\cite{pedersen2017duluth},
\texttt{JU\_CES\_NLP}~\cite{pramanick2017ju},
\texttt{PunFields}~\cite{mikhalkova2017punfields},
\texttt{UWAV}~\cite{vadehra2017uwav},
\texttt{Fermi}~\cite{indurthi2017fermi},
and \texttt{UWaterloo}~\cite{vechtomova2017uwaterloo}. While most of them extract complicated linguistic features to train rule based and machine learning based classifiers.
In addition to task participants,
\texttt{Sense}~\cite{cai2018sense} incorporates word sense representations into RNNs to tackle the homographic pun location task.
The \texttt{CRF}~\cite{zoujoint} captures linguistic features such as POS tags, n-grams, and word suffix  to model puns.
Moreover, the \texttt{Joint}~\cite{zoujoint} jointly models two tasks with RNNs and a CRF tagger.


For the PTD dataset, four baseline methods with reported performance are selected for comparisons.
\texttt{MCL}~\cite{mihalcea2005making} exploits word representations with multiple stylistic features while \texttt{HAE}~\cite{yang2015humor} applies a random forest model with Word2Vec and human-centric features. 
\texttt{PAL}~\cite{chen2017predicting} trains a convolutional neural network (CNN) to learn essential feature automatically. 
Based on existing CNN models, \texttt{HUR}~\cite{chen2018humor} improves the performance  by adjusting the filter size and adding a highway layer.



\begin{table*}[!t]
\centering
\begin{adjustbox}{max width=.9\textwidth}
    \begin{tabular}{|l|ccc|ccc|ccc|ccc|}
        \hline
          \multirow{3}{*}{Model} & \multicolumn{6}{c|}{Homographic Puns} & \multicolumn{6}{c|}{Heterographic Puns} \\
          \cline{2-13}
          & 
          \multicolumn{3}{c|}{Pun Detection} & \multicolumn{3}{c|}{Pun Location} & 
          \multicolumn{3}{c|}{Pun Detection} & \multicolumn{3}{c|}{Pun Location} \\
          \cline{2-13}
          & P & R & \textit{F}$_1$ & P & R & \textit{F}$_1$ & P & R & \textit{F}$_1$ & P & R & \textit{F}$_1$ \\
          \hline
          \hline
          \texttt{Duluth} & 78.32 & 87.24 & 82.54 & 44.00 & 44.00 & 44.00 & 73.99 & 86.62 & 68.71 & - & - & - \\
          \texttt{JU\_CSE\_NLP} & 72.51 & 90.79 & 68.84 & 33.48 & 33.48 & 33.48 & 73.67 & 94.02 & 71.74 & 37.92 & 37.92 & 37.92 \\
          \texttt{PunFields} & 79.93 & 73.37 & 67.82 & 32.79 & 32.79 & 32.79 & 75.80 & 59.40 & 57.47 & 35.01 & 35.01 & 35.01 \\
          \texttt{UWAV} & 68.38 & 47.23 & 46.71 & 34.10 & 34.10 & 34.10 & 65.23 & 41.78 & 42.53 & 42.80 & 42.80 & 42.80 \\
          \texttt{Fermi} & 90.24 & 89.70 & 85.33 & 52.15 & 52.15 & 52.15 & - & - & - & - & - & - \\
          \texttt{UWaterloo} & - & - & - & 65.26 & 65.21 & 65.23 & - & - & - & 79.73 & 79.54 & 79.64 \\
          \texttt{Sense} & - & - & - & 81.50 & 74.70 & 78.00 & - & - & - & - & - & - \\
          \texttt{CRF} & 87.21 & 64.09 & 73.89 & 86.31 & 55.32 & 67.43 & 89.56 & 70.94 & 79.17 & 88.46 & 62.76 & 73.42 \\
          \texttt{Joint} & 91.25 & 93.28 & 92.19 & 83.55 & 77.10 & 80.19 & 86.67 & 93.08 & 89.76 & 81.41 & 77.50 & 79.40 \\
          \hline
          \simplemodelname & 91.42 & 94.21 & 92.79 & 88.80 & 85.65 & 87.20 & 93.35 & 95.04 & 94.19 & 92.31 & 88.24 & 90.23 \\
          \modelname & \textbf{94.18} & \textbf{95.70} & \textbf{94.94} & \textbf{90.43} & \textbf{87.50} & \textbf{88.94} & \textbf{94.84} & \textbf{95.59} & \textbf{95.22} & \textbf{94.23} & \textbf{90.41} & \textbf{92.28}\\
          \hline
    \end{tabular}
    \end{adjustbox}
    \caption{Performance of detecting and locating puns on the SemEval dataset.
    All improvements of \modelname and \simplemodelname over baseline methods are statistically significant at a 95\% confidence level in paired $t$-tests. 
    Comparing to  \modelname, \simplemodelname does not model word pronunciations. Results show that both PCPR and CPR outperform baselines. With modeling pronunciations, PCPR performs the best. 
    }
    \label{tab:results}
\end{table*}

\subsection{Experimental Results}

\noindent \textbf{Pun Detection.}
Table~\ref{tab:results} presents the pun detection performance of methods for both homographic and heterographic puns on the SemEval dataset while Table~\ref{tab:potd} shows the detection performance on the PTD dataset.
For the SemEval dataset, compared to the nine baseline models, \modelname achieves the highest performance with 3.0\% and 6.1\% improvements of \textit{F}$_1$ against the best among the baselines (i.e. \texttt{Joint}) for the homographic and heterographic datasets, respectively.
For the PTD dataset, \modelname improves against \texttt{HUR} by 9.6\%.
Moreover, the variant \simplemodelname beats all of the baseline methods and shows the effectiveness of contextualized word embeddings.
In addition, \modelname further improves the performances by 2.3\% and 1.1\% with the attentive pronunciation feature for detecting homographic and heterographic puns, respectively. 
An interesting observation is that pronunciation embeddings also facilitate homographic pun detection, implying the potential of pronunciation for enhancing general language modeling.

\noindent \textbf{Pun Location.}
Table~\ref{tab:results} shows that the proposed \modelname model achieves highest \textit{F}$_1$-scores on both homographic and heterographic pun location tasks with 10.9\% and 15.9\% incredible increment against the best baseline method.
The improvement is much larger than that on pun detection task.
We posit the reason is that predicting pun locations relies much more on the comparative relations among different tokens in one sentence.
As a result, contextualized word embeddings acquire an enormous advantage.
By applying the pronunciation-attentive representations, different words with similar pronunciations are linked, leading to a much better pinpoint of pun word for the heterographic dataset. 
We notice that some of the baseline models such as \texttt{UWaterloo}, \texttt{UWAV} and \texttt{PunFields} have poor performances. These methods consider the word position in a sentence or calculate the inverse document frequency of words.
We suppose such rule-based recognition techniques can hardly capture the deep semantic and syntactic properties of words.


\begin{table}[!t]
    \centering
    \resizebox{.60\linewidth}{!}{
    \begin{tabular}{|l|ccc|}
    \hline
        Model & P & R & \textit{F}$_1$  \\
        \hline
        \hline
        \texttt{MCL} & 83.80 & 65.50 & 73.50 \\
        \texttt{HAE} & 83.40 & 88.80 & 85.90 \\
        \texttt{PAL} & 86.40 & 85.40 & 85.70 \\
        \texttt{HUR} & 86.60 & 94.00 & 90.10  \\
        \hline
        \simplemodelname & 98.12 & \textbf{99.34} & 98.73 \\
        \modelname & \textbf{98.44} & 99.13 & \textbf{98.79} \\
        \hline
    \end{tabular}
    }
    \caption{Performance of pun detection on the PTD dataset.}    
    \label{tab:potd}
\end{table}

\begin{table}[!t]
    \centering
    \resizebox{\linewidth}{!}{
    \begin{tabular}{|l|ccc|ccc|}
    \hline
      \multirow{2}{*}{Model}  & \multicolumn{3}{c|}{Homographic Puns} & \multicolumn{3}{c|}{Heterographic Puns} \\ \cline{2-7}
      & P & R & \textit{F}$_1$ & P & R & \textit{F}$_1$  \\
      \hline
      \hline
      \texttt{Joint} & 67.70 & 67.70 & 67.70 & 68.84 & 68.84 & 68.84 \\
      \modelname & \textbf{87.21} & \textbf{81.72} & \textbf{84.38} & \textbf{85.16} & \textbf{80.15} & \textbf{82.58} \\
      \hline
    \end{tabular}
    }
    \caption{Performance of pipeline recognition in the SemEval dastaset.}
    \label{tab:pipeline}
\end{table}

\begin{table}[!t]
    \centering
   \resizebox{\linewidth}{!}{
    \begin{tabular}{|l|ccc|}
    \hline
        Model & P & R & \textit{F}$_1$  \\
        \hline
        \hline
        \modelname & \textbf{90.43} & \textbf{87.50} & \textbf{88.94} \\ \hline
        w/o Pre-trained Phoneme Emb. & 89.37 & 85.65 & 87.47  \\
        w/o Self-attention Encoder & 89.17 & 86.42 & 87.70 \\
        w/o Phonological Attention & 89.56 & 87.35 & 88.44  \\
        \hline
    \end{tabular}
    }
    \caption{Ablation study on different features of \modelname for homographic pun detection on the SemEval dataset.}
    \label{tab:ablation}
\end{table}
\begin{figure*}
    \centering
    \includegraphics[width=\linewidth]{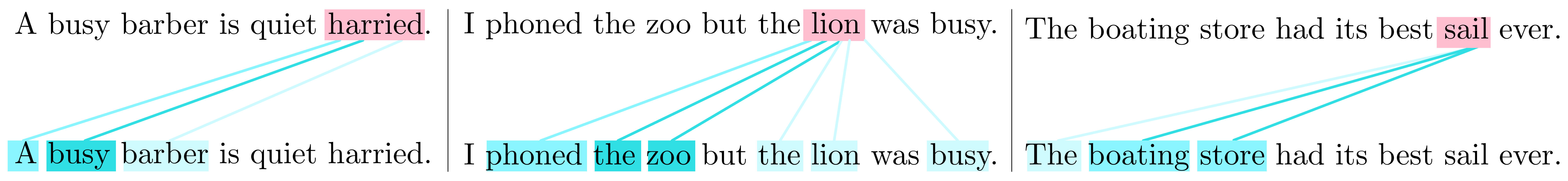}
    \caption{Visualization of attention weights of each pun word (marked in pink) in the sentences. A deeper color indicates a higher attention weight.}
    \label{fig:attention}
\end{figure*}
 
\begin{table*}[!t]
    \centering
   \resizebox{.65\linewidth}{!}{
    \begin{tabular}{|l|c|c|c|}
    \hline
        Sentence & Pun & \simplemodelname & \modelname\\
        \hline
        \hline
        In the dark? Follow the son. & son & - & son \\
        He stole an invention and then told patent lies. & patent & patent & lies \\
        A thief who stole a calendar got twelve months. & got & - & - \\
        \hline
    \end{tabular}
    }
    \caption{A case study of the model predictions for the pun location task of SemEval 2017.}
    \label{tab:case}
\end{table*}

\noindent \textbf{Pipeline Recognition.}
The ultimate goal of pun recognition is to establish a pipeline to detect and then locate puns.
Table~\ref{tab:pipeline} shows the pipeline performances of \modelname and \texttt{Joint}, which is the only baseline with reported pipeline performance for recognizing the homographic and heterographic puns in the SemEval dataset.
\texttt{Joint} achieves suboptimal performance and the authors of \texttt{Joint} attribute the performance drop to error propagation.
In contrast, \modelname improves the \textit{F}$_1$-scores against \texttt{Joint} by 24.6\% and 20.0\% on two pun types.

\subsection{Ablation Study and Analysis}

\noindent \textbf{Ablation Study.}
To better understand the effectiveness of each component in \modelname, we conduct an ablation study on the homographic puns of the SemEval dataset.
Table~\ref{tab:ablation} shows the results on taking out different features of \modelname, including pre-trained phoneme embeddings, the self-attentive encoder, and phonological attention. Note that we use the average pooling as an alternative when we remove the phonological attention module.
As a result, we can see the drop after removing each of the three features.
It shows that all these components are essential for \modelname to recognize puns.

\noindent \textbf{Attentive Weights Interpretation.}
Figure~\ref{fig:attention} illustrates the self-attention weights $\alpha_{i}^S$ of three examples from heterographic puns in the SemEval dataset.
The word highlighted in the upper sentence (marked in pink) is a pun while we also color each word of the lower sentence in blue according to the magnitude of its attention weights.
The deeper colors indicate higher attention weights.
In the first example, \textit{busy} has the largest weight because it has the most similar semantic meaning as \textit{harried}. \textit{The barber} also has relatively high weights. We suppose it is related to \textit{hairy} which should be the other word of this double entendre.
Similar, \textit{the zoo} is corresponded to \textit{lion} while \textit{phone} and \textit{busy} indicate \textit{line} for the pun. Moreover, \textit{boating} confirms \textit{sail} while \textit{store} supports \textit{sale}.  
Interpreting the weights out of our self-attentive encoder explains the significance of each token when the model detects the pun in the context. The phonemes are essential in these cases because they strengthen the relationship among words with distant semantic meanings  but similar phonological expressions.

\begin{figure}[!t]
\centering
    \begin{subfigure}[b]{0.49\linewidth}
        \includegraphics[width=1\linewidth]{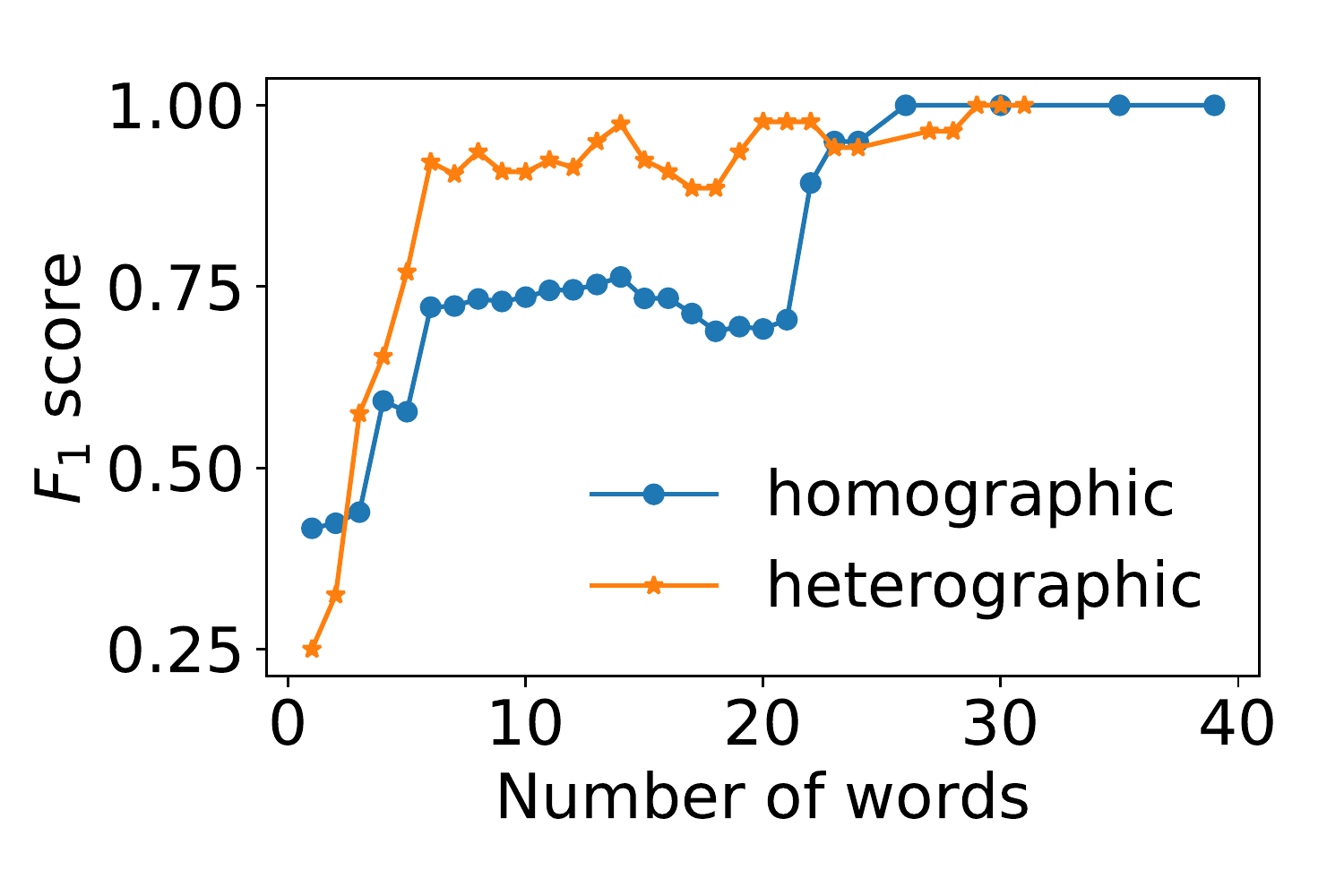}
        \caption{Pun Detection}
        \label{fig:len_de}
    \end{subfigure}
    \begin{subfigure}[b]{0.49\linewidth}
        \includegraphics[width=1\linewidth]{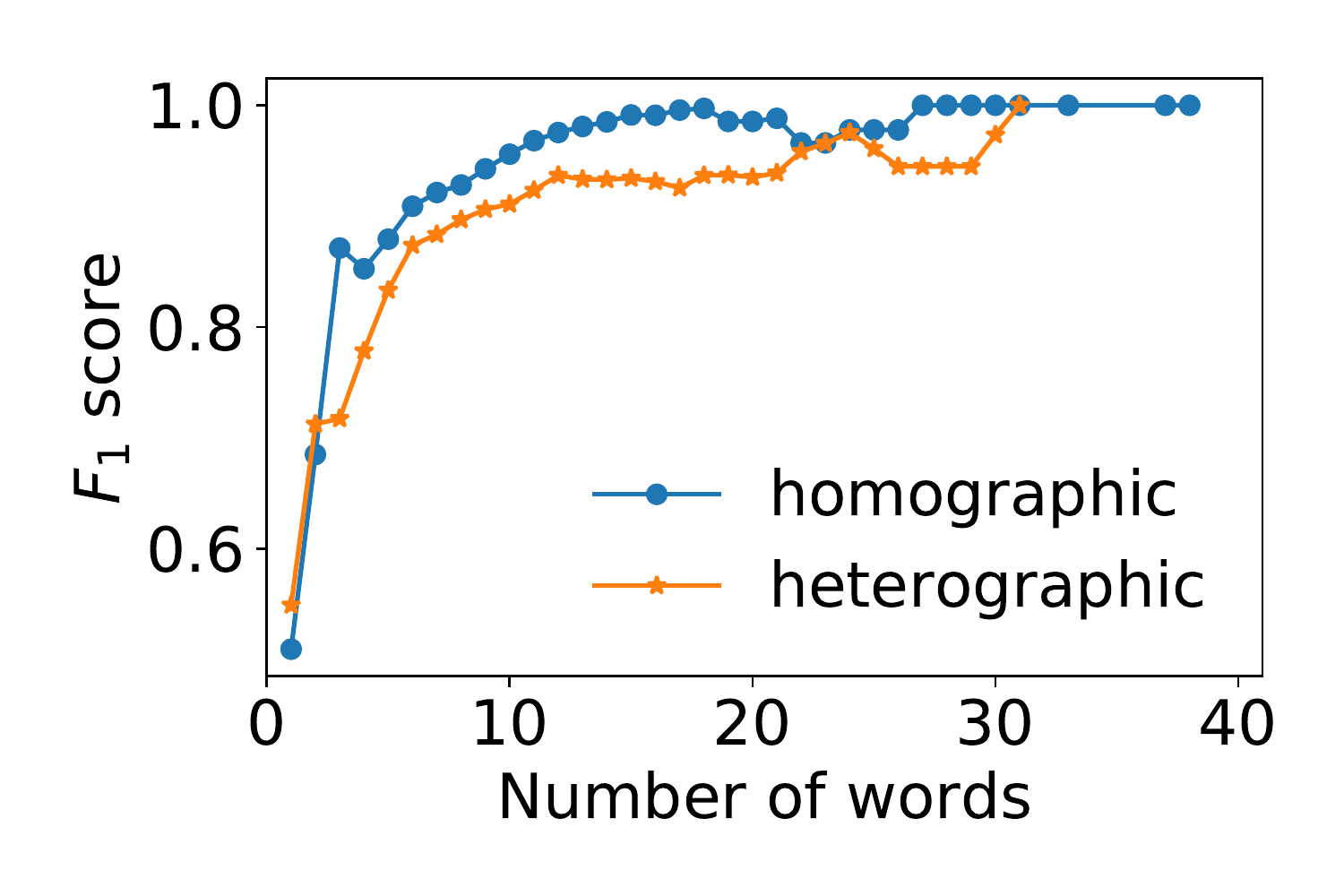}
        \caption{Pun Location}
        \label{fig:len_loc}
    \end{subfigure}
    \caption{
    Pun recognition performance over different text lengths for homographic and heterographic puns on the SemEval dataset.
    }
    \label{fig:len}
\end{figure}

\noindent \textbf{Sensitivity to Text Lengths.} Figure~\ref{fig:len} shows the performance of pun detection and location over different text lengths for homographic and heterographic puns in the SemEval dataset.
For both tasks, the performance gets higher when the text lengths are longer because the context information is richer.
Especially in the pun detection task, we observe that our model requires longer contexts (more than 20 words) to detect the homographic puns. However, shorter contexts (less than 10 words) are adequate for heterographic pun detection, which indicates the contribution from phonological features.
In short, the results verify the importance of contextualized embeddings and pronunciation representations for pun recognition.

\noindent \textbf{Case Study and Error Analysis.}
Table~\ref{tab:case} shows the results of a case study with the outputs of \simplemodelname and \modelname.
In the first case, the heterographic pun comes from the words \textit{son} and \textit{sun}. \simplemodelname fails to recognize the pun word  with limited context information while the phonological attention in \modelname helps to locate it.
However, the pronunciation features in some cases can mislead the model to make wrong predictions. For example, \textit{patent} in the second sentence is a homographic pun word and has several meanings, which can be found with the contextual features.
Besides, the phonemes in \textit{lies} are ubiquitous in many other words like \textit{laws}, thereby confusing the model.
In the last case, \textit{got} is a widely used causative with dozens of meanings so that the word is  hard to be recognized as a pun word with its contextual and phonological features. 

\section{Conclusions}
\label{section:conclusions}

In this paper, we propose a novel approach, \modelname, for pun detection and location by leveraging a contextualized word encoder and modeling phonemes as word pronunciations.
Moreover, we would love to apply the proposed model to other problems, such as general humor recognition, irony discovery, and sarcasm detection, as the future work.

\bibliography{acl2020}

\begin{thebibliography}{53}
\expandafter\ifx\csname natexlab\endcsname\relax\def\natexlab#1{#1}\fi

\bibitem[{Augello et~al.(2008)Augello, Saccone, Gaglio, and
  Pilato}]{augello2008humorist}
Agnese Augello, Gaetano Saccone, Salvatore Gaglio, and Giovanni Pilato. 2008.
\newblock Humorist bot: Bringing computational humour in a chat-bot system.
\newblock In \emph{CISIS 2008}, pages 703--708.

\bibitem[{Bahdanau et~al.(2015)Bahdanau, Cho, and Bengio}]{bahdanau2014neural}
Dzmitry Bahdanau, Kyunghyun Cho, and Yoshua Bengio. 2015.
\newblock Neural machine translation by jointly learning to align and
  translate.
\newblock In \emph{ICLR 2015}.

\bibitem[{Bengio and Heigold(2014)}]{bengio2014word}
Samy Bengio and Georg Heigold. 2014.
\newblock Word embeddings for speech recognition.
\newblock In \emph{ISCA 2014}.

\bibitem[{Bertero and Fung(2016)}]{bertero2016predicting}
Dario Bertero and Pascale Fung. 2016.
\newblock Predicting humor response in dialogues from tv sitcoms.
\newblock In \emph{ICASSP 2016}, pages 5780--5784.

\bibitem[{Blinov et~al.(2019)Blinov, Bolotova-Baranova, and
  Braslavski}]{blinov2019large}
Vladislav Blinov, Valeria Bolotova-Baranova, and Pavel Braslavski. 2019.
\newblock Large dataset and language model fun-tuning for humor recognition.
\newblock In \emph{ACL 2019}, pages 4027--4032.

\bibitem[{Cai et~al.(2018)Cai, Li, and Wan}]{cai2018sense}
Yitao Cai, Yin Li, and Xiaojun Wan. 2018.
\newblock Sense-aware neural models for pun location in texts.
\newblock In \emph{ACL 2018}, pages 546--551.

\bibitem[{Chen and Lee(2017)}]{chen2017predicting}
Lei Chen and Chong~MIn Lee. 2017.
\newblock Predicting audience's laughter using convolutional neural network.
\newblock \emph{arXiv preprint arXiv:1702.02584}.

\bibitem[{Chen and Soo(2018)}]{chen2018humor}
Peng-Yu Chen and Von-Wun Soo. 2018.
\newblock Humor recognition using deep learning.
\newblock In \emph{NAACL 2018}, pages 113--117.

\bibitem[{Devlin et~al.(2018)Devlin, Chang, Lee, and
  Toutanova}]{devlin2018bert}
Jacob Devlin, Ming-Wei Chang, Kenton Lee, and Kristina Toutanova. 2018.
\newblock Bert: Pre-training of deep bidirectional transformers for language
  understanding.
\newblock \emph{arXiv preprint arXiv:1810.04805}.

\bibitem[{Diao et~al.(2018)Diao, Lin, Wu, Yang, Xu, Yang, Wang, Zhang, Xu, and
  Zhang}]{diao2018weca}
Yufeng Diao, Hongfei Lin, Di~Wu, Liang Yang, Kan Xu, Zhihao Yang, Jian Wang,
  Shaowu Zhang, Bo~Xu, and Dongyu Zhang. 2018.
\newblock Weca: A wordnet-encoded collocation-attention network for homographic
  pun recognition.
\newblock In \emph{EMNLP 2018}, pages 2507--2516.

\bibitem[{Doogan et~al.(2017)Doogan, Ghosh, Chen, and Veale}]{doogan2017idiom}
Samuel Doogan, Aniruddha Ghosh, Hanyang Chen, and Tony Veale. 2017.
\newblock Idiom savant at semeval-2017 task 7: Detection and interpretation of
  english puns.
\newblock In \emph{SemEval-2017}, pages 103--108.

\bibitem[{Ghosh et~al.(2015)Ghosh, Li, Veale, Rosso, Shutova, Barnden, and
  Reyes}]{ghosh2015semeval}
Aniruddha Ghosh, Guofu Li, Tony Veale, Paolo Rosso, Ekaterina Shutova, John
  Barnden, and Antonio Reyes. 2015.
\newblock Semeval-2015 task 11: Sentiment analysis of figurative language in
  twitter.
\newblock In \emph{SemEval 2015}, pages 470--478.

\bibitem[{He et~al.(2019)He, Peng, and Liang}]{he2019pun}
He~He, Nanyun Peng, and Percy Liang. 2019.
\newblock Pun generation with surprise.
\newblock In \emph{NAACL 2019}.

\bibitem[{Hurtado et~al.(2017)Hurtado, Segarra, Pla, Carrasco, and
  Gonz{\'a}lez}]{hurtado2017elirf}
Llu{\'\i}s-F Hurtado, Encarna Segarra, Ferran Pla, Pascual Carrasco, and
  Jos{\'e}-Angel Gonz{\'a}lez. 2017.
\newblock Elirf-upv at semeval-2017 task 7: Pun detection and interpretation.
\newblock In \emph{SemEval-2017}, pages 440--443.

\bibitem[{Indurthi and Oota(2017)}]{indurthi2017fermi}
Vijayasaradhi Indurthi and Subba~Reddy Oota. 2017.
\newblock Fermi at semeval-2017 task 7: Detection and interpretation of
  homographic puns in english language.
\newblock In \emph{SemEval-2017}, pages 457--460.

\bibitem[{Jaech et~al.(2016)Jaech, Koncel-Kedziorski, and
  Ostendorf}]{jaech2016phonological}
Aaron Jaech, Rik Koncel-Kedziorski, and Mari Ostendorf. 2016.
\newblock Phonological pun-derstanding.
\newblock In \emph{ACL 2016}, pages 654--663.

\bibitem[{Kamper et~al.(2016)Kamper, Wang, and Livescu}]{kamper2016deep}
Herman Kamper, Weiran Wang, and Karen Livescu. 2016.
\newblock Deep convolutional acoustic word embeddings using word-pair side
  information.
\newblock In \emph{ICASSP 2016}, pages 4950--4954. IEEE.

\bibitem[{Kingma and Ba(2014)}]{kingma2014adam}
Diederik~P Kingma and Jimmy Ba. 2014.
\newblock Adam: A method for stochastic optimization.
\newblock \emph{arXiv preprint arXiv:1412.6980}.

\bibitem[{LeCun et~al.(1995)LeCun, Bengio et~al.}]{lecun1995convolutional}
Yann LeCun, Yoshua Bengio, et~al. 1995.
\newblock Convolutional networks for images, speech, and time series.
\newblock \emph{The handbook of brain theory and neural networks},
  3361(10):1995.

\bibitem[{Lee et~al.(2019)Lee, Yoon, Kim, Kim, Kim, So, and
  Kang}]{lee2019biobert}
Jinhyuk Lee, Wonjin Yoon, Sungdong Kim, Donghyeon Kim, Sunkyu Kim, Chan~Ho So,
  and Jaewoo Kang. 2019.
\newblock Biobert: pre-trained biomedical language representation model for
  biomedical text mining.
\newblock \emph{arXiv preprint arXiv:1901.08746}.

\bibitem[{Luo et~al.(2019)Luo, Li, Yang, Li, Chang, Sui, and
  SUN}]{luo-etal-2019-pun}
Fuli Luo, Shunyao Li, Pengcheng Yang, Lei Li, Baobao Chang, Zhifang Sui, and
  Xu~SUN. 2019.
\newblock Pun-{GAN}: Generative adversarial network for pun generation.
\newblock In \emph{Proceedings of the 2019 Conference on Empirical Methods in
  Natural Language Processing and the 9th International Joint Conference on
  Natural Language Processing (EMNLP-IJCNLP)}.

\bibitem[{McCann et~al.(2017)McCann, Bradbury, Xiong, and
  Socher}]{mccann2017learned}
Bryan McCann, James Bradbury, Caiming Xiong, and Richard Socher. 2017.
\newblock Learned in translation: Contextualized word vectors.
\newblock In \emph{NeurIPS 2017}, pages 6294--6305.

\bibitem[{Melby and Warner(1995)}]{melby1995possibility}
Alan~K Melby and Terry Warner. 1995.
\newblock \emph{The possibility of language: A discussion of the nature of
  language, with implications for human and machine translation}, volume~14.
\newblock John Benjamins Publishing.

\bibitem[{Mihalcea and Strapparava(2005)}]{mihalcea2005making}
Rada Mihalcea and Carlo Strapparava. 2005.
\newblock Making computers laugh: Investigations in automatic humor
  recognition.
\newblock In \emph{EMNLP 2005}, pages 531--538.

\bibitem[{Mikhalkova and Karyakin(2017)}]{mikhalkova2017punfields}
Elena Mikhalkova and Yuri Karyakin. 2017.
\newblock Punfields at semeval-2017 task 7: Employing roget's thesaurus in
  automatic pun recognition and interpretation.
\newblock \emph{arXiv preprint arXiv:1707.05479}.

\bibitem[{Mikolov et~al.(2018)Mikolov, Grave, Bojanowski, Puhrsch, and
  Joulin}]{mikolov2018advances}
Tomas Mikolov, Edouard Grave, Piotr Bojanowski, Christian Puhrsch, and Armand
  Joulin. 2018.
\newblock Advances in pre-training distributed word representations.
\newblock In \emph{LREC 2018}.

\bibitem[{Mikolov et~al.(2013)Mikolov, Sutskever, Chen, Corrado, and
  Dean}]{mikolov2013distributed}
Tomas Mikolov, Ilya Sutskever, Kai Chen, Greg~S Corrado, and Jeff Dean. 2013.
\newblock Distributed representations of words and phrases and their
  compositionality.
\newblock In \emph{Advances in neural information processing systems}, pages
  3111--3119.

\bibitem[{Miller(1998{\natexlab{a}})}]{miller1998pronunciation}
Corey~Andrew Miller. 1998{\natexlab{a}}.
\newblock Pronunciation modeling in speech synthesis.

\bibitem[{Miller(1998{\natexlab{b}})}]{miller1998wordnet}
George Miller. 1998{\natexlab{b}}.
\newblock \emph{WordNet: An electronic lexical database}.
\newblock MIT press.

\bibitem[{Miller et~al.(2017)Miller, Hempelmann, and
  Gurevych}]{miller2017semeval}
Tristan Miller, Christian Hempelmann, and Iryna Gurevych. 2017.
\newblock Semeval-2017 task 7: Detection and interpretation of english puns.
\newblock In \emph{SemEval-2017}, pages 58--68.

\bibitem[{Navigli(2009)}]{navigli2009word}
Roberto Navigli. 2009.
\newblock Word sense disambiguation: A survey.
\newblock \emph{ACM computing surveys (CSUR)}, 41(2):10.

\bibitem[{Nijholt et~al.(2017)Nijholt, Niculescu, Alessandro, and
  Banchs}]{nijholt2017humor}
Anton Nijholt, Andreea~I Niculescu, Valitutti Alessandro, and Rafael~E Banchs.
  2017.
\newblock Humor in human-computer interaction: a short survey.

\bibitem[{Oele and Evang(2017)}]{oele2017buzzsaw}
Dieke Oele and Kilian Evang. 2017.
\newblock Buzzsaw at semeval-2017 task 7: Global vs. local context for
  interpreting and locating homographic english puns with sense embeddings.
\newblock In \emph{SemEval-2017}, pages 444--448.

\bibitem[{Pedersen(2017)}]{pedersen2017duluth}
Ted Pedersen. 2017.
\newblock Duluth at semeval-2017 task 7: Puns upon a midnight dreary, lexical
  semantics for the weak and weary.
\newblock \emph{arXiv preprint arXiv:1704.08388}.

\bibitem[{Pennington et~al.(2014)Pennington, Socher, and
  Manning}]{pennington2014glove}
Jeffrey Pennington, Richard Socher, and Christopher Manning. 2014.
\newblock Glove: Global vectors for word representation.
\newblock In \emph{EMNLP 2014}, pages 1532--1543.

\bibitem[{Peters et~al.(2018)Peters, Neumann, Iyyer, Gardner, Clark, Lee, and
  Zettlemoyer}]{peters2018deep}
Matthew Peters, Mark Neumann, Mohit Iyyer, Matt Gardner, Christopher Clark,
  Kenton Lee, and Luke Zettlemoyer. 2018.
\newblock Deep contextualized word representations.
\newblock In \emph{NAACL 2018}, pages 2227--2237.

\bibitem[{Peters et~al.(2017)Peters, Ammar, Bhagavatula, and
  Power}]{peters2017semi}
Matthew~E Peters, Waleed Ammar, Chandra Bhagavatula, and Russell Power. 2017.
\newblock Semi-supervised sequence tagging with bidirectional language models.
\newblock \emph{arXiv preprint arXiv:1705.00108}.

\bibitem[{Powers(2011)}]{powers2011evaluation}
David~Martin Powers. 2011.
\newblock Evaluation: from precision, recall and f-measure to roc,
  informedness, markedness and correlation.

\bibitem[{Pramanick and Das(2017)}]{pramanick2017ju}
Aniket Pramanick and Dipankar Das. 2017.
\newblock Ju cse nlp $@ $ semeval 2017 task 7: Employing rules to detect and
  interpret english puns.
\newblock In \emph{SemEval-2017}, pages 432--435.

\bibitem[{Schuster and Paliwal(1997)}]{schuster1997bidirectional}
Mike Schuster and Kuldip~K Paliwal. 1997.
\newblock Bidirectional recurrent neural networks.
\newblock \emph{IEEE Transactions on Signal Processing}, 45(11):2673--2681.

\bibitem[{Sch{\"u}tze et~al.(2007)Sch{\"u}tze, Manning, and
  Raghavan}]{schutze2007introduction}
Hinrich Sch{\"u}tze, Christopher~D Manning, and Prabhakar Raghavan. 2007.
\newblock \emph{An introduction to information retrieval}.
\newblock Cambridge University Press,.

\bibitem[{Sukhbaatar et~al.(2015)Sukhbaatar, Weston, Fergus
  et~al.}]{sukhbaatar2015end}
Sainbayar Sukhbaatar, Jason Weston, Rob Fergus, et~al. 2015.
\newblock End-to-end memory networks.
\newblock In \emph{Advances in neural information processing systems}, pages
  2440--2448.

\bibitem[{Toutanova and Moore(2002)}]{toutanova2002pronunciation}
Kristina Toutanova and Robert~C Moore. 2002.
\newblock Pronunciation modeling for improved spelling correction.

\bibitem[{Vadehra(2017)}]{vadehra2017uwav}
Ankit Vadehra. 2017.
\newblock Uwav at semeval-2017 task 7: Automated feature-based system for
  locating puns.
\newblock In \emph{SemEval-2017}, pages 449--452.

\bibitem[{Vaswani et~al.(2017)Vaswani, Shazeer, Parmar, Uszkoreit, Jones,
  Gomez, Kaiser, and Polosukhin}]{vaswani2017attention}
Ashish Vaswani, Noam Shazeer, Niki Parmar, Jakob Uszkoreit, Llion Jones,
  Aidan~N Gomez, {\L}ukasz Kaiser, and Illia Polosukhin. 2017.
\newblock Attention is all you need.
\newblock In \emph{Advances in neural information processing systems}, pages
  5998--6008.

\bibitem[{Vechtomova(2017)}]{vechtomova2017uwaterloo}
Olga Vechtomova. 2017.
\newblock Uwaterloo at semeval-2017 task 7: Locating the pun using syntactic
  characteristics and corpus-based metrics.
\newblock In \emph{SemEval-2017}, pages 421--425.

\bibitem[{Wu et~al.(2016)Wu, Schuster, Chen, Le, Norouzi, Macherey, Krikun,
  Cao, Gao, Macherey et~al.}]{wu2016google}
Yonghui Wu, Mike Schuster, Zhifeng Chen, Quoc~V Le, Mohammad Norouzi, Wolfgang
  Macherey, Maxim Krikun, Yuan Cao, Qin Gao, Klaus Macherey, et~al. 2016.
\newblock Google's neural machine translation system: Bridging the gap between
  human and machine translation.
\newblock \emph{arXiv preprint arXiv:1609.08144}.

\bibitem[{Xiu et~al.(2017)Xiu, Lan, and Wu}]{xiu2017ecnu}
Yuhuan Xiu, Man Lan, and Yuanbin Wu. 2017.
\newblock Ecnu at semeval-2017 task 7: Using supervised and unsupervised
  methods to detect and locate english puns.
\newblock In \emph{SemEval-2017}, pages 453--456.

\bibitem[{Yang et~al.(2015)Yang, Lavie, Dyer, and Hovy}]{yang2015humor}
Diyi Yang, Alon Lavie, Chris Dyer, and Eduard Hovy. 2015.
\newblock Humor recognition and humor anchor extraction.
\newblock In \emph{EMNLP 2015}, pages 2367--2376.

\bibitem[{Yang et~al.(2019)Yang, Dai, Yang, Carbonell, Salakhutdinov, and
  Le}]{yang2019xlnet}
Zhilin Yang, Zihang Dai, Yiming Yang, Jaime Carbonell, Ruslan Salakhutdinov,
  and Quoc~V Le. 2019.
\newblock Xlnet: Generalized autoregressive pretraining for language
  understanding.
\newblock \emph{arXiv preprint arXiv:1906.08237}.

\bibitem[{Yu et~al.(2018)Yu, Tan, and Wan}]{yu2018a}
Zhiwei Yu, Jiwei Tan, and Xiaojun Wan. 2018.
\newblock A neural approach to pun generation.
\newblock In \emph{ACL 2018}.

\bibitem[{Zhu et~al.(2018)Zhu, Jin, Ni, Wei, and Lu}]{zhu2018improve}
Wenhao Zhu, Xin Jin, Jianyue Ni, Baogang Wei, and Zhiguo Lu. 2018.
\newblock Improve word embedding using both writing and pronunciation.
\newblock \emph{PloS one}, 13(12):e0208785.

\bibitem[{Zou and Lu(2019)}]{zoujoint}
Yanyan Zou and Wei Lu. 2019.
\newblock Joint detection and location of english puns.
\newblock In \emph{NAACL 2019}, pages 2117--2123.

\end{thebibliography}
\bibliographystyle{acl_natbib}

\end{document}